# A review of neuro-fuzzy systems based on intelligent control

**Fatemeh Zahedi**[*], **Zahra Zahedi**

Electrical and Computer Engineering, Shiraz University, Shiraz, Iran

**Email address:**

Fatemeh.zahedi28@gmail.com (F. Zahedi), Zahra.zahedi28@gmail.com (Z. Zahedi)

**Abstract:** The system's ability to adapt and self-organize are two key factors when it comes to how well the system can survive the changes to the environment and the plant they work within. Intelligent control improves these two factors in controllers. Considering the increasing complexity of dynamic systems along with their need for feedback controls, using more complicated controls has become necessary and intelligent control can be a suitable response to this necessity. This paper briefly describes the structure of intelligent control and provides a review on fuzzy logic and neural networks which are some of the base methods for intelligent control. The different aspects of these two methods are then compared together and an example of a combined method is presented.

## 1. Introduction

In the last few decades, fuzzy logic has been identified as one of the most active and useful fields of research and study, and has found many uses in many different applications. Of its most important usages are non-linear control systems, time variant systems, and control systems whose dynamics are exactly known, such as servo-motor position control systems and robot arm control systems.

In fuzzy logic systems, decisions are made based on the inputs received in the form of linguistic variables. These variables are grouped up in fuzzy sets and are assigned a degree of membership within that set which are determined by certain formulas that are referred to as membership functions. Fuzzy logic rules are defined as preconditions in the form of if-then rules defined based on the fuzzy sets and the response of each rule is determined through fuzzy implication.

Currently, there are no systematic methods for designing systems based on fuzzy logic. The easiest and the fastest method is to define membership functions and fuzzy rules based on human-operated systems which are then tested until the desired output is obtained [1].

Current research aims at designing systems that work on fuzzy logic and learn from their own experiences. However, there are few articles available that discuss the creating of fuzzy control rules and improvement of those rules from experience. Some of the presented methods can have hopeful results but these methods are mainly subjective and heuristic and finding the proper membership functions are done through trial and error.

On the other hand, applying the neural network's ability to learn, to fuzzy logic systems provides us with more methods that can yield hopeful results. These methods which are referred to as neuro-fuzzy methods create a desirable framework for finding solutions for more complicated problems. If the available knowledge can be expressed in the form of linguistic rules we can have a fuzzy logic system. And if we have the necessary data and are able to learn from that data through simulations, we can use neural networks. Specifying the fuzzy sets, the fuzzy operators and the knowledge base are requirements for building a fuzzy logic system, and specifying the architecture and learning algorithms are requirements for creating a neural network [2].

In the final section of this paper, the cooperative neuro-fuzzy model is reviewed. In this method, the parameters, including the fuzzy rules, rule weights and fuzzy sets are determined using neural networks.

## 2. Intelligent Control

Generally, modeling most complicated processes can't be achieved by mathematical function or simple physical rules. Intelligent control aims at finding a solution to this problem by combining intelligent and creative characteristics of human controllers. Intelligent control performs under an environment with lots of uncertainties and unexpected situations in a way that keeps its integrity and consistency with that environment and rectifies failures in the system without limits.

Intelligent controls, regardless of what structure or configuration they have, should have the following attributes [3]:

A. Correctness: the ability to perform a specific set of needed operations.

B. Robustness: the ability of the system to maintain its functions under unexpected and unusual situations that can occur internally or externally.

C. Extensibility: the ability of the intelligent control to support planned and unplanned upgrades (hardware or software) without the need for re-design.

D. Reusability & compatibility: the ability to use sub-systems and components in different applications and compatibility with new situations.

There are three basic methods for intelligent control two of which are fuzzy logic and neural networks.

## 3. Neural network

Control engineers consider neural networks to be large scale, non-linear dynamic systems that are defined within a first order differential equation. Neural networks are in fact new structures for information processing systems that consist of numerous linked processing elements. The links that connect these processing elements are called interconnections. "Fig. 1" is an example of a single processing element. Each processing element has one or more inputs and a single output. Every input has a weight assigned to it and these weights vectors change by learning rules.

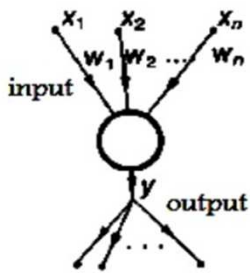

*Figure 1. A sample processing element*

There can be a question of how such a large dynamic system processes its information? The answer to this question is the use of energy functions for the system. Each non-linear dynamic system has several equilibrium points. These points are the minimum points on the energy landscape. If an arbitrary input pattern is given to the system as its initial state, the system is capable of approaching one of these equilibrium points subject to the global stability of the system. For example, according to "Fig. 2", E(X) is an energy function with three equilibrium points of P1, P2 and P3. If X(0) is the initial state of the system, X(0) should converge to the closest equilibrium point.

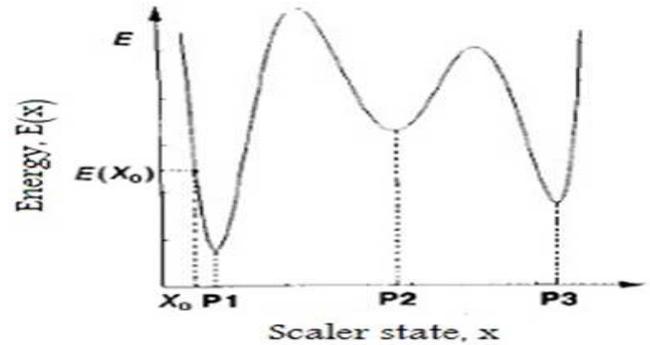

*Figure 2. Energy landscape with 3 minimum points*

For the defined system in the differential "(1)" and "(2)", the Lyapunov function (energy) would be as the "(3)".

$$\dot{x}_i = a_i(x_i)[b_i(x_i) - \sum_{k=1}^{n} w_{ik} d_k(x_k)] \quad (1)$$

$$\dot{w}_i = w_{ij} + G(w_{ij}, x_i, x_j) \quad (2)$$

$$V(x) = -\sum_{i=1}^{n} \int_0^{x_i} b_i(\varepsilon_i) d'_i(\varepsilon_i) d\varepsilon_i + \tfrac{1}{2} \sum_{j,k=1}^{n} w_{jk} d_j(x_j) d_k(x_k) \quad (3)$$

## 4. Fuzzy Logic

### 4.1. Defining some Fuzzy Operations

We start out by defining some fuzzy operations. If A and B are two fuzzy sets in the universal set of U with membership functions of $\mu_A$ and $\mu_B$, then:

Definition 1- Union: The membership function $\mu_{(A \cup B)}$ of the union A∪B for all u∈U is defined as followed:

$$\mu_{(A \cup B)} = \max \{\mu_A(u), \mu_B(u)\} \quad (4)$$

Definition 2- Intersection: The membership function $\mu_{(A \cap B)}$ of the intersection $A \cap B$ for all u∈U is defined as followed:

$$\mu_{(A \cap B)} = \min \{\mu_A(u), \mu_B(u)\} \quad (5)$$

Definition 3- Complement: The membership function $\mu_{\bar{A}}$ of the complement A for all u∈U is defined as followed:

$$\mu_{\bar{A}}(u) = 1 - \mu_A(u) \quad (6)$$

### 4.2. Architecture of Fuzzy Logic Controllers

"Fig. 3" shows the basic configuration of fuzzy logic controllers which is consisted of four parts.

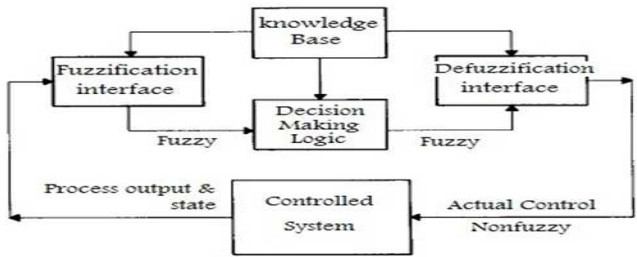

*Figure 3. Basic configuration of fuzzy logic controllers*

1- Fuzzification interface which is consisted of the following steps:
- Measuring the value of input variables.
- Transfer of a wide range of input variable values to the related universal set.
- Fuzzification, which is in fact the conversation of input data to proper linguistic value.

2- Knowledge base, including "data base" and "linguistic (fuzzy) control rule base".
- The data base provides the necessary definitions that are used to define linguistic control rules and fuzzy data.
- The rule base which is specifies the control goals using a set of linguistic control rules.

3- Decision making logic, which is the kernel of the fuzzy logic controller. In this part human decision making is simulated based on fuzzy concepts and inferring fuzzy control actions by using fuzzy implications and rules of inference.

4- Defuzzification interface which is consisted of the following steps:
- Conversion of a wide range of output variables to universal set.
- Deffuzification of rules and linguistic variables in a way that is retrievable and understandable by the rest of process.

## 5. Neuro-Fuzzy

### 5.1. Neuro-Fuzzy Structure

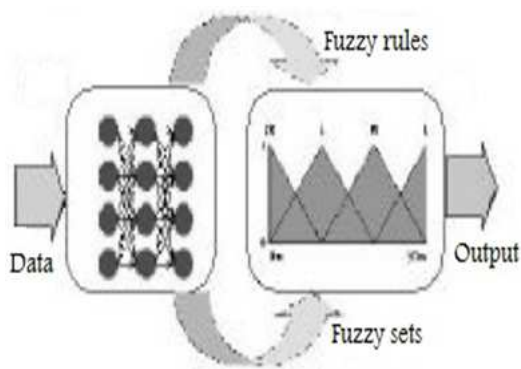

*Figure 4. Neuro-fuzzy cooperative model*

Neuro-fuzzy is a model in which the neural network uses the training data to determine the membership functions and fuzzy rules of the fuzzy logic system. After specifying the parameters of the fuzzy logic system, the neural network moves to the margins. Basic rules are usually determined using fuzzy clustering algorithms. The neural network approximates the membership functions from the training data. "Fig. 4" shows a neuro-fuzzy cooperative model.

An example of a neuro-fuzzy structure is as follows [1], [2]:

As it can be seen in "Fig. 5", this structure contains five layers. Two linguistic nodes are present for each output, one for the training data (the desired output), and another for the actual output of the neuro-fuzzy system.

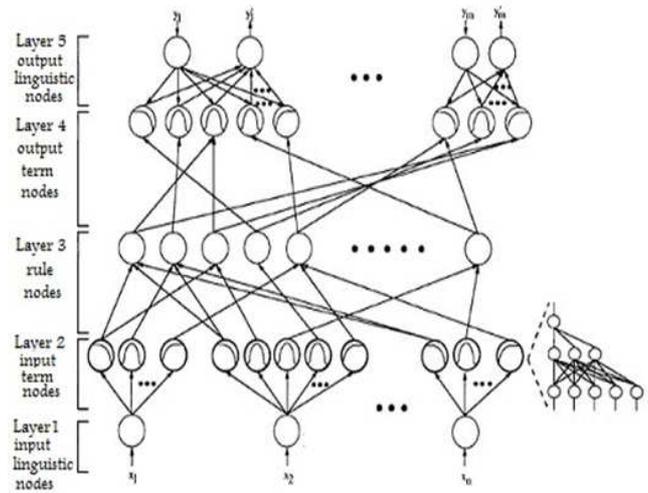

*Figure 5. The five layer neuro-fuzzy structure*

The first internal or hidden layer is in charge of the fuzzification of the input variables. Each node in this layer can be either singular which identifies a simple membership function, or it can be consisted of multi-layered nodes which calculate the complex membership functions.

The second hidden layer defines the preconditions and rules resulted from the third hidden layer. In this method, the output is generated using the hybrid-learning algorithm consisted of unsupervised learning to retrieve the initial membership functions and rule base.

### 5.2. Fuzzy Control of an Unmanned Vehicle

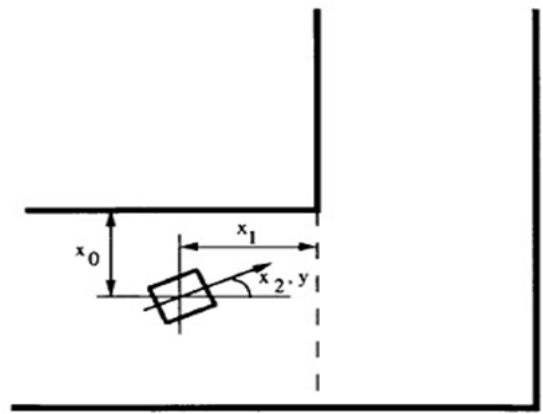

*Figure 6. The state variables of the fuzzy vehicle*

In this section, an example is presented to show the main application of this model [1]. The vehicle uses examples to learn to automatically move through a track with rectangular turns. $X_0$, $X_1$ and $X_2$ are the linguistic variables that specify the distance from track boundaries from one side and the current angle of the steering wheel. The output linguistic variable Y is the next steering angle. This concept is better explained in "Fig. 6".

"Fig. 7", shows the membership functions of the $x_0$, $x_1$, $x_2$ and y variables, learned after one or two training phases.

"Table, 1" shows the learned fuzzy logic rules. These rules show the hidden layers of the neural network design. In this example, we assume that the vehicle moves at a constant speed and sensors are installed on the vehicle for measuring the x0, x1 and x2 variables. These variables are given to the controller so that it can derive the next steering angle from them. This example has been simulated under different initial conditions for the steering angle and the results have been very close to reality.

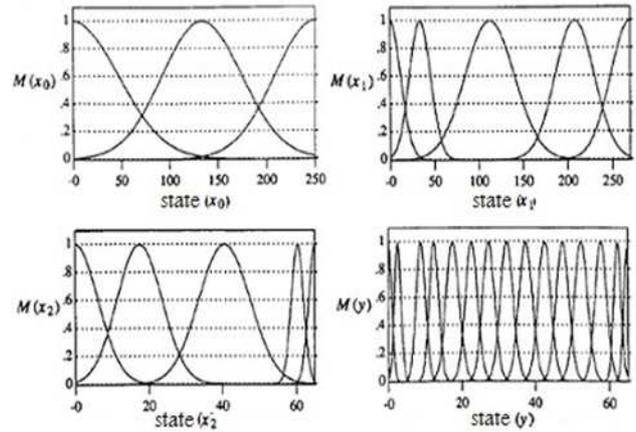

*Figure 7.* The learned membership functions

*Table 1.* Learned fuzzy logic rules in this example

| Learned fuzzy logic rules | | | | | | | | | |
|---|---|---|---|---|---|---|---|---|---|
| Rule | Precondition | | | Consequence | Rule | Precondition | | | consequence |
| | $x_0$ | $x_1$ | $x_2$ | Y | | $x_0$ | $x_1$ | $x_2$ | Y |
| 0 | 0 | 1 | 0 | 6 | 15 | 1 | 2 | 1 | 4 |
| 1 | 1 | 1 | 0 | 6 | 16 | 2 | 2 | 1 | 2 |
| 2 | 2 | 1 | 0 | 5 | 17 | 0 | 3 | 1 | 2 |
| 3 | 0 | 2 | 0 | 3 | 18 | 1 | 3 | 1 | 2 |
| 4 | 1 | 2 | 0 | 2 | 19 | 2 | 3 | 1 | 2 |
| 5 | 2 | 2 | 0 | 1 | 20 | 0 | 0 | 2 | 11 |
| 6 | 0 | 3 | 0 | 2 | 21 | 1 | 0 | 2 | 10 |
| 7 | 1 | 3 | 0 | 2 | 22 | 2 | 0 | 2 | 12 |
| 8 | 2 | 3 | 0 | 0 | 23 | 0 | 1 | 2 | 10 |
| 9 | 0 | 0 | 1 | 9 | 24 | 1 | 1 | 2 | 8 |
| 10 | 1 | 0 | 1 | 9 | 25 | 2 | 1 | 2 | 13 |
| 11 | 0 | 1 | 1 | 7 | 26 | 0 | 2 | 2 | 7 |
| 12 | 1 | 1 | 1 | 7 | 27 | 1 | 2 | 2 | 7 |
| 13 | 2 | 1 | 1 | 7 | 28 | 2 | 2 | 2 | 13 |
| 14 | 0 | 2 | 1 | 6 | 29 | 0 | 3 | 2 | 13 |

## 6. Discussion and Conclusion

In this paper, intelligent control and two of its basic methods, neural networks and fuzzy logic were generally represented. Cooperative neuro-fuzzy model was reviewed and the general structure of one of its methods was explained. An example was provided for the two main applications of the model, fuzzy logic controllers and fuzzy decision making systems. This example shows the superiority of the hybrid learning method over the more traditional methods.